\documentclass[10pt,twocolumn,letterpaper]{article}

\usepackage{cvpr}
\usepackage{times}
\usepackage{epsfig}
\usepackage{graphicx}
\usepackage{amsmath}
\usepackage{amssymb}
\usepackage{bm,array,multirow}
\usepackage[breaklinks=true,bookmarks=false]{hyperref}

\cvprfinalcopy % *** Uncomment this line for the final submission

\def\cvprPaperID{****} % *** Enter the CVPR Paper ID here

\ifcvprfinal\fi
\begin{document}

%%%%%%%%% TITLE
\title{Deeply Matting-based Dual Generative Adversarial Network for Image and Document Label Supervision}

\author{Yubao Liu$^{*}$
~~~Kai Lin$^{*}$
}

\maketitle
\thispagestyle{empty}

%%%%%%%%% ABSTRACT
\begin{abstract}
Although
many methods have been proposed
to deal with nature image super-resolution (SR)
and get impressive performance,
the text images SR is not good
due to
their ignorance of 
document images.
In this paper,
we propose a
matting-based dual generative adversarial network (mdGAN)
for document image SR.
Firstly,
the input image is decomposed into document
text, foreground and background layers
using deep image matting.
Then
two parallel branches are constructed to
recover
text boundary information and color information respectively.
Furthermore,
in order
to improve the 
restoration accuracy of characters in output image, 
we use
the input image's corresponding 
ground truth text label  
as extra supervise information 
to refine the two-branch networks during training.
Experiments on real text images
demonstrate that our method outperforms several state-of-the-art methods quantitatively and qualitatively.
\end{abstract}

%%%%%%%%% BODY TEXT
\section{Introduction}

Single image super-resolution (SR) 
aims to magnify 
low-resolution (LR)
image to high resolution (HR) one.
Generally speaking,
there are three kind of methods
for image SR:
interpolation based methods \cite{anbarjafari2010image,dai2009softcuts},
reconstruction based methods \cite{kim2010single,yang2010image}
and exemplar training \cite{Freeman2002Example,Kim2008Example} based methods.
However,
the SR problem is ill-posed 
due to a multiplicity of 
solutions exist for any given LR pixel.
Recently,
deep neural networks \cite{dong2014learning,lim2017enhanced,wang2018recovering,pan2018learning}
improve the performance significantly
in terms of quantitatively and qualitatively.
However,
the performance of these networks has decreased 
toward text images 
due to their not taking the special properties of 
text images.
Different from natural images,
text images have more high-frequency information.
In this paper,
we focus on text image SR.

Text image SR aims to recover HR image  
with sharp text boundaries.
Traditional methods \cite{Walha2012Super,Agam2013A}
add extra restrictions while solving an
ill-posed problem.
Nowadays,
learning-based methods 
have witnessed great performance in low-level vision tasks.
ASRS \cite{walha2015resolution}
restores textual images via 
multiple coupled dictionaries 
and adaptive sparse representation selection.
In \cite{Dong2015Boosting},
CNN is proposed for text image SR 
and achieves remarkable performance.
SRCNN \cite{dong2014learning} is used to 
improve the accuracy of OCR
by pre-processing.
In \cite{Pandey2017Language},
language independent text SR is constructed 
using CNN
and it takes less time during test
than most natural images SR.
Unfortunately,
the text boundaries in output HR images 
of these methods
are not sharp enough and are with some blur.
Besides,
some characters on the HR images
may suffer from deformation
when the resolution of LR images is very low.

To address these issues,
we propose a matting-based dual generative adversarial network (mdGAN)
for text image SR.
The contributions of this paper are as follows:
First,
a deep matting method is used to 
decompose the input text image 
into two color images(foreground and background) and a text boundary image. 
Then two parallel branches 
are constructed to restore the 
two color images and the text boundary image respectively.
The advantage is that
it helps reduce blurring artifacts
by processing the color and text boundary separately.
Second,
the LR image's
corresponding text label
is used as extra supervise information
to reduce characters deformation in output HR image.
To our best knowledge,
we are the first one 
to introduce text label 
to improve the quality of restoration
for image SR.

The rest of this paper is organized as follows. The related work is presented in Section 2,
followed by the details of our method in Section 3.
Section 4 presents the experimental results. 
The conclusion is drawn in Section 5.

\section{Related work}

\subsection{Image matting}
Image matting is to estimate
foreground color layer, background color layer and alpha matte
layer of an image.
Many methods 
\cite{Shahrian2013Improving,Cho2016Natural,Shen2016Deep}
may suffer from 
high-frequency chunky and low-frequency smearing artifacts
due to 
highly reliant on color-dependent propagation.
Instead of relying primarily on color information,
Xu et al. \cite{Xu2017Deep} 
learn the natural structure that is presented in alpha mattes.
An encoder-decoder
is constructed  
to
takes image and the corresponding trimap as input
and predicts the alpha matte.
Then they use a network to refine the alpha matte.

\subsection{Generative adversarial networks}

%GAN%

GANs have achieved great success 
in many 
low-level vision tasks,
including image denoising,
image deblurring
and image super-resolution \cite{wang2018recovering,Ledig2017Photo,wang2018esrgan}.
Despite their success,
the training of GANs
is known to be unstable
and has great influence on performance.
Many works \cite{Radford2015Unsupervised,Zhang2016StackGAN,Karras2017Progressive} have been proposed to
stabilize the GAN's training dynamics and improve the sample's diversity.

\section{The proposed method}
The architecture of our proposed  
matting-based dual generative adversarial network (mdGAN)
is shown in Fig. 2. 
It mainly consists of three steps:
(1) decomposes the input LR text image 
into two color layers (foreground and background) and one text boundary layer
using a deep matting method;
(2) recover each layer separately using two parallel
branches
and compose the outputs to generate the output HR image;
(3) calculate character classification loss in HR image 
based on an extra text label  
and adopt the loss
to refine the two-branch networks
in step 2.

\subsection{Matting-based image decomposition}
To reduce blurring artifacts
in the process of text image SR,
we introduce image matting in our network.
Image matting 
claims that an image
can be decomposed into 
a foreground layer, a background layer and 
an alpha matte layer.
In this paper,
alpha matte is called definite
text boundary layer.
Foreground and background layers
are called foreground and background color layers respectively.
An example of three layers are shown
in Fig. 3.
The text boundary layer contains
the soft edges of text.
The foreground and background layers
only contain the color information
of the text and background.
By processing the color and text boundary layers
respectively,
it contributes to reducing 
blurring artifacts for 
text image SR.
In this paper,
deep image matting \cite{Xu2017Deep}
is used in our matting process.

\subsection{Parallel SR branches process stage}

As shown in Fig. 2,
text boundary layer $\alpha^{L}$
contains the major text shape information
while
foreground layer $F^{L}$ and background layer $B^{L}$
are smooth but have more color information.
Therefore,
we adopt two parallel SR branches 
to 
recover 
the three layers respectively.
For foreground and background layers,
ESRGAN \cite{wang2018esrgan} is used as branch-1
due to its stronger supervision for color consistency.
Owing to strong ability of restoration 
for high-frequency details,
SFTGAN \cite{wang2018sftgan}
is adopted to 
recover the text boundary layer $\alpha^{L}$
as the branch-2.
To highlight the boundary information
of $\alpha^{L}$,
teager filtering \cite{Mancas2005Super} is adopted
to $\alpha^{L}$ before the $\alpha^{L}$ is given to the branch-2.
Thanks to the fact that we have used image matting to separate the input LR images into 
color information and boundary information,
the filtering won't influence
the color information of the original 
image.
Therefore,
our method can avoid
artifacts like color aliasing.
Besides, 
to stable the training of GANs
and improve the quality of restoration,
we apply spectral normalization(SN) \cite{miyato2018spectral} to 
parameters in the branch-1 and branch-2.
Once the three HR layers
$F^{H}$, $B^{H}$ and $\alpha^{H}$ are obtained,
the output HR image can be calculated as follows: 
\begin{equation}
I^{H}=
\alpha^{H}F^{H}+
(1-\alpha^{H})B^{H}
\end{equation}

%%%fig ori

\section{Experimental results}

In this section,
we
demonstrate 
our text image SR performance
by conducting experiments
on 
one published text image SR dataset
and one dataset constructed by ourselves.
Besides,
ablation study is
carried out 
to demonstrate the effectiveness of
each part of our architecture.

\subsection{Qualitative evaluation}
We evaluate the performance of our methods against several state-of-the-art image SR methods: bicubic, SRCNN \cite{dong2014learning},
SRGAN \cite{Ledig2017Photo},
EDSR \cite{lim2017enhanced},
ESRGAN \cite{wang2018esrgan}
and 
RCAN \cite{Zhang2018Image}.
All models of these methods
are from scratch trained
under the training datasets
which are adopted in our model.
For ICDAR 2015,
RMSE, PSNR and SSIM results are shown in Table 1.
Owing to the dataset is used to
improve the OCR accuracy,
thus the OCR accuracy after SR
is also displayed in Table 1. 
We can see that our method outperforms the other methods 
with a large margin.
Fig. 1 and Fig. 4 show the visually comparison results.
Our method products  
sharper text boundaries and less artifacts.
For super TextSR 2018,
the quantitative and qualitative comparisons
are shown in Table 2 and Fig. 5.
%ablation study to switch 
\subsection{Ablation study}
To evaluate the effectiveness of every part of
our architecture,
we measure quantitative results
under
baseline,
with image matting,
with character classifier loss
and with both image matting and 
character classifier loss
in Table 3.
From Table 3 we can see that
the quantitative results
keep increasing with the matting and 
the character classification loss,
which demonstrates 
the effectiveness of each part.

%ICDAR 2015 TEXTSR%
\begin{table}[!t]
\scriptsize
\renewcommand\arraystretch{0.9}
\centering
\caption{Results of different methods on the ICDAR
2015 TEXTSR dataset}
\tabcolsep=4pt
\begin{tabular}{|p{1.3cm}<{\centering}|p{1.3cm}<{\centering}|p{1.3cm}<{\centering}|p{1.3cm}<{\centering}|p{1.3cm}<{\centering}|}
\hline
 Method & RMSE & PSNR & MSSIM & OCR(\%)  \\
\hline
 Bicubic & 19.04 & 23.50 & 0.879 & 60.64  \\
 \hline
 SRCNN & 7.24 & 33.19 & 0.981 & 76.10  \\
 \hline
 SRGAN & 7.10 & 33.51 & 0.987 & 76.80  \\
 \hline
 EDSR  & 7.02 & 33.60 & 0.990 & 77.13   \\
 \hline
 ESRGAN & 6.92 & 33.68 & 0.992 & 77.59   \\
 \hline
 RCAN & 6.86 & 33.71 & 0.994 & 77.64   \\
 \hline
 Ours & 6.80 & \textbf{33.90} & \textbf{0.996} & \textbf{77.78}  \\
 \hline
\end{tabular}
\label{tab:example}
\end{table}

\section{Conclusion}

In this paper,
we propose mdGAN for text image SR.
To reduce blurring artifacts in output HR image,
the input LR image is firstly
decomposed into 
color and text boundary layers 
by deep image matting.
These layers  
are then processed SR by
two parallel branches separately.
The outputs of the two branches
are composed
to generate 
the output HR image.
To reduce character deformation in output HR image,
a character classification loss 
and a deep similar loss
are used to refine the two branches.
Besides, spectral normalization is adopted in two branches
to improve the recover quality.
Extensive experiments demonstrate that
our method
outperforms several state-of-the-art methods
quantitatively and qualitatively.

% % BibTeX References
% {\small
% \bibliographystyle{ieee}
% \bibliography{icme2019template}
% }

\end{document}